# Data-Driven Portfolio Management for Motion Pictures Industry: A New Data-Driven Optimization Methodology Using a Large Language Model as the Expert

Mohammad Alipour-Vaezi, Kwok-Leung Tsui[*]

*Grado Department of Industrial & Systems Engineering, Virginia Tech, Blacksburg, VA 24061, USA*

**Abstract:** Portfolio management is one of the unresponded problems of the Motion Pictures Industry (MPI). To design an optimal portfolio for an MPI distributor, it is essential to predict the box office of each project. Moreover, for an accurate box office prediction, it is critical to consider the effect of the celebrities involved in each MPI project, which was impossible with any precedent expert-based method. Additionally, the asymmetric characteristic of MPI data decreases the performance of any predictive algorithm. In this paper, firstly, the fame score of the celebrities is determined using a large language model. Then, to tackle the asymmetric character of MPI's data, projects are classified. Furthermore, the box office prediction takes place for each class of projects. Finally, using a hybrid multi-attribute decision-making technique, the preferability of each project for the distributor is calculated, and benefiting from a bi-objective optimization model, the optimal portfolio is designed.

**Declarations of Interest:** None

**Funding Source:** None

---

[*] Corresponding Author
Email Addresses: Alipourvaezi@vt.edu (M. Alipour-Vaezi); kltsui@vt.edu (K.L. Tsui)
ORCID IDs: 0000-0002-7529-1848 (M. Alipour-Vaezi); 0000-0002-0558-2279 (K.L. Tsui)

# 1. Introduction

The effective management of properties has emerged as a fundamental capability for thriving enterprises, particularly those operating in the domains of entertainment and creativity (Nurimbetov & Metyakubov, 2020). Presently, culture and art constitute vital components of every society, alongside other developmental facets, contributing significantly to the growth and advancement of nations. The Motion Pictures Industry (MPI), which entertains millions of people around the globe, is a multi-billion-dollar industry that employs thousands of people in various roles. Only in 2019, the global box office revenue was $42.5 billion, and the industry supported 2.5 million jobs worldwide (Mcclintock, 2020).

The greater the size of an industry, the more crucial it becomes to get optimized in different aspects. Many researchers addressed managerial problems of the Media Supply Chain (MSC), including the MPI to optimize various problems it encounters, such as scheduling advertisements (Alipour-Vaezi, Tavakkoli-Moghaddam, et al., 2022), predicting cinema-going behavior (Jones et al., 2007), and designing a movie recommender system based on audience preferences (Vellaichamy & Kalimuthu, 2017).

One of the major problems in the MSC that has been neglected in the precedent research studies is to optimize the portfolio of MPI's distributors. Portfolio management, which can be defined as the process of selecting and managing a collection of financial assets (Elton et al., 2009), is a crucial aspect of strategic decision-making (Martinsuo & Vuorinen, 2023). Each organization seeks to select the most efficient and most profitable project among available projects (Song et al., 2023). That is to say, the goal of Portfolio Management is to maximize profits toward the company's strategic goals (Jang & Seong, 2023). To do this, managers need to identify, select, prioritize, and manage an appropriate portfolio of projects (Bilgin et al., 2017).

MPI, which involves the production, distribution, and exhibition of movies, has been subject to numerous changes and challenges in recent years, including the rise of digital technologies, shifts in consumer preferences, and the impact of the COVID-19 pandemic (Sokowati, 2022; Yaqoub et al., 2023). To navigate this dynamic landscape, investment companies operating in this field must adopt effective portfolio management strategies that can help them maximize their returns while mitigating risks. For this reason, audience preferences have to be predicted, which cannot be reached unless using Machine Learning (ML) algorithms that discover the knowledge within the historical data. Also, from the point of view of an MPI's distributor, it is necessary to consider the rate of return of each project alongside various other criteria that the distributor may have to optimize the portfolio in question.

This research project aims to provide a comprehensive and practical framework for effective portfolio management in the MPI supply chain, helping MPI's investment companies navigate this complex landscape with confidence and success. The proposed methodology fills the gap existing in this field using a portfolio optimization model benefitting from ML models and Multi-Attribute Decision-Making (MADM) techniques.

The rest of this article is organized as follows: In Section 2, the problem under study will be described and the assumptions considered will be illuminated. Later, in the same section, the steps of the proposed data-driven method for optimizing the MPI's distributor portfolio are described. Section 3 is dedicated to the primary analysis and data-mining process. In Section 4, the method proposed for predicting the box office is verified. Moreover, in Section 5, the proposed model is evaluated benefiting from some numerical experiments. Also, in Section 6, the sensitivity of the proposed model is analyzed. Finally, in Section 7, the research conclusion and limitations are counted down alongside some suggestions for future research.

**2. Theoretical Framework & Methodology**

It is critical for distributors to make sure the projects they are investing in are profitable. It is important to note that, in the majority of cases, such as those involving MPI, the task of predicting the profitability of a project can be challenging due to the interdependence of multiple factors within the field (Hennig-Thurau et al., 2007).

One of the factors that can influence the profitability of an MPI project is the celebrities who are taking part in it (e.g., actors, writers, and directors). Celebrities have a considerable influence on consumers' purchase decisions (Fleck et al., 2012). Obviously, each celebrity's presence in an MPI project can act as an incentive for the audience and lead to higher profit for MPI distributors.

Even if the profitability of the projects is guaranteed, the distributors still need to decide which of the projects should be considered in their portfolio. However, profitability is not the only key factor for designing an MPI portfolio. Each distributor has a personal preference based on previous experience or their expectations.

According to the above-mentioned problem described, the following research question will arise:

- How the profitability of an MPI project can be predicted?
- How the influence of celebrities on the profitability of an MPI project can be measured?
- How can the distributor design a portfolio of MPI projects effectively?
- How the distributor's preference can be considered in the decision-making process?

To answer these questions, this paper proposes a novel data-driven optimization methodology for portfolio management of the MPI's distributors. A few assumptions have been taken into account, which can be listed as follows:

- There is no uncertainty, and all the parameters are discrete numbers.

- Throughout the study, there would be no change in the distributor's budget or project features.
- Distributor has no other motivation than profitability for investing in a project.

Figure 1 illustrates the steps of the proposed methodology. The subsequent sections and paragraphs will thoroughly and individually cover each step.

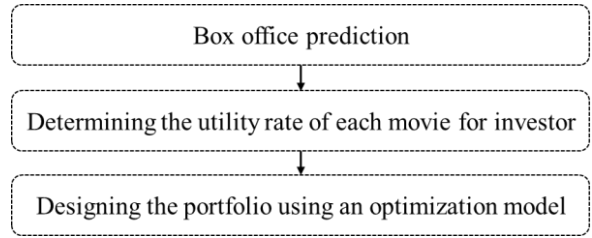

*Figure 1. Schematic overview of the proposed methodology.*

## 2.1. Box Office Prediction

To design the optimal portfolio for MPI's distributors, it is essential to know the profit each project will bring to them. In this research, benefitting from the best-fitted ML model, the box office amount of each MPI's project will be predicted. Figure 2 depicts the steps needed to be taken to predict the box office of each project. Also, each step is discussed in the following subsections. It's a delight to mention that the preprocessed dataset, classification codes, and regression codes have been provided as Supplementary Materials I-III.

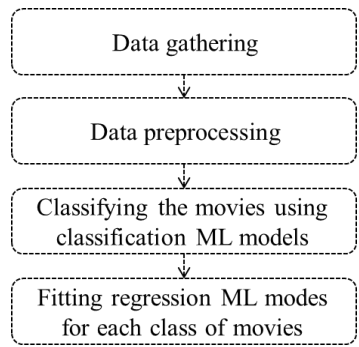

*Figure 2. Box office prediction steps.*

### 2.1.1. Data Gathering

Gathering the dataset is a crucial aspect of this study, as it enables the examination of the insights it holds. Hence, it is necessary to obtain a fitting dataset that can be utilized to train and test the ML models. For this reason, a comprehensive dataset has been collected regarding the movies released in the United States. This data set covers 6097 movies from 1980 to 2020. Table 1 displays the features of the gathered dataset.

*Table 1. Data Description.*

| Feature | Type |
|---|---|
| Movie's Rating | Categorical |
| Movie's Genre | Categorical |
| Franchise or Sequel | Boolean |
| Released Month | Numerical |
| IMDb Score | Numerical |
| IMDb Vote Counts | Numerical |
| Director's Fame Score | Numerical |
| Writer's Fame Score | Numerical |
| Lead Actor/Actress Fame Score | Numerical |
| Domesticity | Boolean |
| Run Time | Numerical |
| Budget | Numerical |
| Box Office | Numerical |

*2.1.2. Data Preprocessing*

For ML models to generate accurate and dependable results, they necessitate high-quality input data. In some cases, data may contain anomalies that analysts must identify and overcome. By employing Microsoft Office Excel alongside Python programming language to execute pre-processing procedures, we produced a train and test dataset containing 6028 records. These pre-processing measures included eliminating or modifying missing or incomplete data (Lai et al., 2019), normalization of both discrete and continuous data (Alipour-Vaezi, Tavakkoli-Moghadaam, et al., 2022), the conversion of qualitative data into quantitative data (Mamabolo & Myres, 2019), the balancing of data (Batista et al., 2004), and the extraction of important features (Mutlag et al., 2020).

It is worth mentioning that we employed the one-hot encoding technique (Yu et al., 2022) to handle categorical features, such as 'Movie's Rating' and 'Movie's Genre.' Additionally, we calculated the current value of financial-related features, such as 'Budget' and 'Box Office,' by adjusting for the inflation rate of the United States. To detect and eliminate the animalities, we used the Isolation Forest ML algorithm (Liu et al., 2008). Also, for feature dimension reduction, we used the Principal Component Analysis (PCA) Algorithm (Karamizadeh et al., 2013).

One of the key factors that determines the success or failure of MPI's projects is the celebrities involved, given that this industry is highly reliant on people and their brands (Baber & Fanea-Ivanovici, 2021; De Vany & Walls, 2004; Tasker, 1993). Specifically, the audience may be more inclined to watch a movie if a renowned celebrity is involved, particularly in a leading role such as a director or lead actor/actress (Barbas, 2016; Luo et al., 2010).

In this study, we sought to examine the impact of celebrities on a movie's box office performance by using three features: 'Director's Fame Score,' 'Writer's Fame Score,' and 'Lead Actor/Actress Fame Score.' These scores represent the notability of each of these individuals, which, as previously stated, can have a significant influence on a movie's box office success.

One effective method for computing these fame scores is to solicit feedback from an expert or panel of experts. Nevertheless, as previously noted, our dataset encompasses a portion of movies released in the U.S. in 40 years between 1980 and 2020, and it may be challenging to find individuals possessing such extensive knowledge of MPI. Moreover, relying on expert-based judgments poses the risk of incorporating their convictions, ideologies, and backgrounds, which could potentially influence their feedback (Wilcock, 1985). Also, even if an expert is found with this vast knowledge, no one can determine the fame scores only based on the celebrity's performance before the movie's release time without considering their activities up to now.

Based on the aforementioned concerns, we utilized Chat Generative Pre-Trained Transformer (Chat GPT), a Large Language Model (LLM) that has undergone training in multiple domains and is presently gaining widespread adoption. Chat GPT can be perceived as being more knowledgeable than a human expert while lacking a personal background that could bias its opinions. As a preliminary step, we provided Chat GPT with the names of the director, writer, and lead actor/actress for each movie in our dataset, along with the year of release, and obtained its assessment of their fame score. These scores are distributed from 0 to 10.

For train-test splitting, this study used a 30% ratio (70% for training and 30% for testing). It is worth mentioning that the splitting is conducted by benefiting from the cross-validation technique (Fushiki, 2011).

2.1.3. Classifying the Movies Using Classification ML Models

MPI datasets are asymmetric in their character. One movie can successfully produce billions of dollars, while another may receive less than a thousand dollars. Addressing this issue can be approached through various means. One option involves the removal of outliers, aiming to enhance data integrity. Alternatively, a stratified approach can be adopted wherein akin projects are categorized, and the learning process is executed independently for each segment. However, it is imperative to acknowledge that the former method may entail the elimination of a substantial portion of data, while the latter can be implemented through the utilization of either classification or clustering machine learning algorithms.

Classification and clustering ML algorithms serve different purposes. Classification involves training a model to assign predefined labels to new, unseen data based on its features (Kumar & Verma, 2012), while clustering involves grouping data points based on inherent similarities without predefined labels, aiming to discover patterns or structures within the data (Xu & Tian, 2015). The advantages of classification over clustering include clear interpretability, as each class has a distinct meaning, making it easier to understand and communicate results. Additionally, classification provides a structured framework for prediction and decision-making, allowing for the automation of tasks like spam filtering or image recognition. In contrast, clustering aims to group similar data points without predefined labels, making it useful for exploratory data analysis but lacking the explicit utility of assigning meaningful categories to new instances. Classification is advantageous when distinct, predefined categories are essential for decision-making and interpretation.

Accordingly, this research applies Classification ML Algorithms to classify the movies into three main classes based on their box office range. Table 2 provides more detail about the classes considered. It should be mentioned that these thresholds are selected by an MPI expert considering the distribution of the MPI projects.

*Table 2. Classes description.*

| Class Number | Box Office Range | Distribution |
|---|---|---|
| 1 | Box Office < 100,000,000 | 3891 |
| 2 | 100,000,000 ≤ Box Office < 1,000,000,000 | 2024 |
| 3 | 1,000,000,000 ≤ Box Office | 182 |

As it is shown in Table 2, class 3 only comprises 182 MPI projects (2.9% of the dataset). According to the huge difference between projects, this class is studied within this research.

The classification algorithms used in this research include Random Forest, Extra Trees, Stochastic Gradient Boosting, XG Boost, Light Gradient Boosting, Cat Boost, Decision Tree, Support Vector Machine, Logistic Regression, and K-Nearest Neighbors. Also, using the Voting Ensemble algorithm, all other ten ML models mentioned earlier, are utilized as one classification algorithm.

To measure the performance of the above-mentioned classification algorithms, we use a few performance measurements, including accuracy, precision, sensitivity (Recall), F-1 score, and Matthew's Correlation Coefficient (MCC). Equations (1-5) illustrate the mathematical formulation for each of these performance measures. Also, for better understanding, Table 3 provides the notations used for these formulations (Alipour-Vaezi, Tavakkoli-Moghadaam, et al., 2022).

*Table 3. Notations used for classification algorithms' performance measures.*

| Parameter | Notation | Description |
|---|---|---|
| True Positive | $TP$ | Positive records that are correctly predicted by the model. |
| True Negative | $TN$ | Negative records that are correctly predicted by the model. |

| False Positive | FP | Positive records that are falsely predicted by the model. |
| False Negative | FN | Negative records that are falsely predicted by the model. |

$$\text{Accuracy} = \frac{TP + TN}{TP + FP + FN + TN} \tag{1}$$

$$\text{Precision} = \frac{TP}{TP + FP} \tag{2}$$

$$\text{Recall} = \frac{TP}{TP + FN} \tag{3}$$

$$\text{F1 Score} = \frac{2 * TP}{(2 * TP) + FP + FN} \tag{4}$$

$$\text{MCC} = \frac{(TP * TN) - (FP * FN)}{\sqrt{(TP + FP)(TP + FN)(TN + FP)(TN + FN)}} \tag{5}$$

*2.1.4. Fitting Regression ML Models for Each Class of Movies*

This section aims to develop the ability to predict the box office performance of the prospective movies in each class. Using the gathered dataset, regression ML algorithms are employed to forecast the potential revenue of each upcoming movie. The goal is to gain a better understanding of the factors that influence the box office success of movies and to improve the accuracy of revenue forecasting for the MPI.

The ML algorithms used in this section of the research include Linear Regression (Su et al., 2012), Ridge Regression (McDonald, 2009), LASSO Regression (Ranstam & Cook, 2018), Decision Tree Regression (Xu et al., 2005), Random Forest Regression (Rodriguez-Galiano et al., 2015), Gradient Boosting Regression (Yang et al., 2020), and XG Boost Regression (Zhang et al., 2020).

This study evaluates the performance of the above-mentioned regression ML models fitted on the gathered dataset based on the following measures: Mean Absolute Error (MAE), Mean Squared Error (MSE), Root Mean Squared Error (RMSE), Mean Absolute Percentage Error (MAPE), and Coefficient of determination ($R^2$). Equations (6-10) mathematically depict these measures (Chicco et al., 2021; Willmott & Matsuura, 2005). Where, $y_i$, $\hat{y}_i$, and $\bar{y}_i$ represent the actual value of the i[th] observation, predicted value of the i[th] observation, and the average value of all the actual observations. Also, $n$ shows the number of observations.

$$MAE = \frac{1}{n} \sum_{i=1}^{n} |y_i - \hat{y}_i| \tag{6}$$

$$MSE = \frac{1}{n} \sum_{i=1}^{n} (y_i - \hat{y}_i)^2 \tag{7}$$

$$RMSE = \sqrt{\frac{1}{n}\sum_{i=1}^{n}(y_i - \hat{y}_i)^2} \tag{8}$$

$$MAPE = \frac{1}{n}\sum_{i=1}^{n}\left|\frac{y_i - \hat{y}_i}{y_i}\right| \tag{9}$$

$$R^2 = 1 - \frac{\sum_{i=1}^{n}(y_i - \hat{y}_i)^2}{\sum_{i=1}^{n}(y_i - \bar{y}_i)^2} \tag{10}$$

## 2.2. Determining the Preferability Rate of Each Movie for Distributor

Similar to any other field, MPI's distributors have some personal preferences regarding the projects they are willing to consider in their portfolio. In this research, in order to calculate the preferability of each project for distributors, a combination of the Bayesian Best-Worst Method (BBWM) and the Weighted Aggregated Sum Product Assessment method (WASPAS) is used. Figure 3 identifies the steps of the proposed methodology for calculating the preferability rate of each movie for distributors.

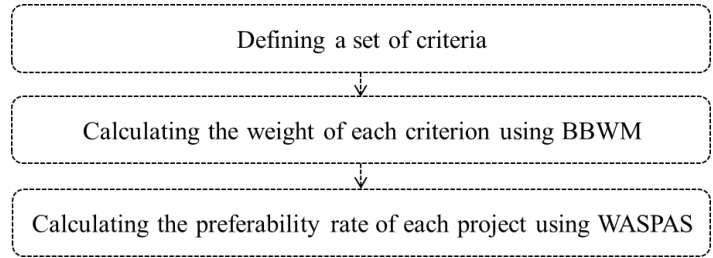

*Figure 3. Steps of preferability rate calculation using MADM techniques*

### 2.2.1. Defining a Set of Criteria

The first step for calculating the preferability rate of each project for the distributor is to define a set of criteria. For this reason, a group of 3 experts working in the field of MPI has been established. Then, a set of criteria has been defined using the Delphi technique (Okoli & Pawlowski, 2004). Table 4 lists the criteria set and their descriptions.

*Table 4. Criteria description*

| Criterion | Notation | Description |
|---|---|---|
| Movie's Rating | $c_1$ | The movie's rating is a qualitative assessment that reflects the content, and appeal of the film helping viewers gauge its potential enjoyment or suitability. |
| Movie's Genre | $c_2$ | A movie's genre categorizes it based on its thematic elements, narrative style, and tone, helping audiences anticipate the type of story and |
| | | experience they can expect. |
| Director's Fame Score | $c_3$ | The Director's Fame Score quantifies the recognition and influence of a filmmaker based on factors like past successes, critical acclaim, and industry reputation. |
| Writer's Fame Score | $c_4$ | The Writer's Fame Score measures the prominence and impact of a screenplay writer considering their previous works, recognition, and impact on the industry. |
| Lead Actor/Actress Fame Score | $c_5$ | The Lead Actor/Actress's Fame Score evaluates the visibility and significance of a performer in the entertainment world, considering their past roles, popularity, and influence. |

### 2.2.2. Calculating the weight of Each Criterion using BBWM

The Bayesian Best-Worst Method (BBWM) is a recently developed but efficient MADM technique proposed by Mohammadi and Rezaei (2020). In the following paragraphs, the steps of this technique will be discussed.

It is essential to note that this MADM technique is an extension of the Best-Worst Method (BWM) proposed by Rezaei (2015), which is shown in Model (11). The initial phase in BWM involves the establishment of a delineated set of criteria earmarked for the subsequent analysis and comparative assessment. Subsequently, the following procedural phase entails the identification of the foremost pivotal criteria, as well as the identification of the least significant criteria within the predefined set. After the delineation of the best and the worst criteria, the expert is tasked with the allocation of preference values to the best criterion concerning the remaining criteria ($A_B = (a_{B,1}, a_{B,2}, \dots, a_{B,n})$). This is achieved through the utilization of a numerical scale ranging from 1 (denoting equivalent importance) to 9 (indicating markedly greater significance). Similarly, the assessment of preferences for each criterion in comparison to the least significant (worst) criterion is conducted ($A_W = (a_{1,W}, a_{2,W}, \dots, a_{n,W})^T$). Employing model (11) and utilizing the vectors $A_B$ and $A_W$, the optimal weight vector is computed ($w^* = (w_1^*, w_2^*, \dots, w_n^*)$).

$$\min_{w} \max_{j} \left\{ \left| \frac{w_B}{w_j} - a_{B,j} \right|, \left| \frac{w_j}{w_W} - a_{j,W} \right| \right\}$$

S.t.

$$\sum_{j=1}^{n} w_j = 1$$

$$w_j \geq 0 \qquad \forall j$$

(11)

Assume that there is a group of $K$ experts ($k = 1, ..., K$) who formed a set of criteria ($c_1, ..., c_n$) and identified the best ($c_B$) and the worst ($c_W$) criteria. Each expert evaluates each remaining criterion, comparing to the best and the worst criteria, which will by each expert. This evaluation will form the vectors $A_B^{1:K}$ and $A_W^{1:K}$.

The objective of the BBWM model is to optimize the overall optimal weight ($w^{agg}$). Using the vector of optimal weights of the set of criteria determined by each expert ($w^k$), $w^{agg}$ can be calculated. It is noteworthy that the joint probability distribution can be computed by applying Equation (12).

$$P(w^{agg}, w^{1:k} | A_B^{1:k}, A_W^{1:k}) \tag{12}$$

As per the probability distribution established by Equation (12), one can calculate the probability of each variable using Equation (13). It is important to note that in Equation (13), $x$ and $y$ refer to any two random variables.

$$P(X) = \sum_y P(x, y) \tag{13}$$

In Equation (14), the Bayes rule is shown for this problem. Furthermore, Equation (15) applies the Bayes rule to take into account the independence among various variables that exist within the joint probability.

$$P(A_W^K | w^{agg}, w^k) = P(A_W^k | w^k) \tag{14}$$

$$P(w^{agg}, w^{1:K} | A_B^{1:K}, A_W^{1:K}) \propto P(A_B^{1:K}, A_W^{1:K} | w^{agg}, W^{1:K}) P(w^{agg}, w^{1:K})$$
$$= P(w^{agg}) \prod_{k=1}^{K} P(A_W^k | w^k) P(A_B^k | w^k) P(w^k | w^{agg}) \tag{15}$$

The $A_B$ and $A_W$ can be effectively represented by the multinomial distribution, which preserves the fundamental concept of BWM. Although $A_B$ and $A_W$ are different in that $A_W$ indicates the preference for all the criteria over the worst, while $A_B$ represents the preference of the best overall the other criteria. As such, they can be modeled according to the approach outlined in Equations (16-17).

$$A_B^k | w^k \sim multinomial\left(\frac{1}{w^k}\right) \tag{16}$$

$$A_W^k | w^k \sim multinomial(w^k) \tag{17}$$

Once $w^{agg}$ is determined, each $w^k$ is in close proximity to it. In Equation (18), the Dirichlet distribution has been reparametrized based on its mean and concentration parameters.

$$w^k | w^{agg} \sim Dir(y \times w^{agg}) \qquad \forall k = 1,2, \ldots, K \tag{18}$$

Condition (18) stipulates that $w^{agg}$ represents the mean of the distribution, whereas $\gamma$ signifies the concentration parameter. According to Equation (18), each weight vector $w^k$ associated with an expert must be located in the vicinity of $w^{agg}$, given that it serves as the mean of the distribution. The proximity between these weight vectors is governed by the non-negative parameter γ. Furthermore, the concentration parameter also requires modeling with a distribution. A suitable choice would be the gamma distribution, which satisfies the non-negativity condition outlined in Equation (19).

$$y = gamma(a, b) \tag{19}$$

Condition (18) depicts $a$ and $b$ as the shape parameters of the gamma distribution. By utilizing the BBWM model, one can ultimately furnish the prior distribution over $w^{agg}$ through an uninformative Dirichlet distribution, represented as Relation (20) with the parameter $\alpha$ set to 1.

$$w^{agg} \sim Dir(\alpha) \tag{20}$$

### 2.2.3. Calculating the Preferability Rate of Each project using WASPAS

The WASPAS proposed by Zavadskas et al. (2012) is an effective MADM technique for ranking the alternatives. The WASPAS method can be considered as a hybrid of two well-known MADM models: The Weighted Sum Model (WSM) and the Weighted Product Model (WPM), as presented below:

The initial phase of the WASPAS technique involves normalizing the decision matrix. For positive (beneficial) and negative (non-beneficial) criteria, this is accomplished using Equation (21).

$$\bar{x}_{ij} = \begin{cases} \dfrac{x_{ij}}{\min\limits_{i} x_{ij}} & \text{For beneficial criteria} \\ \dfrac{\min\limits_{i} x_{ij}}{x_{ij}} & \text{For non-beneficial criteria} \end{cases} \tag{21}$$

Where $x_{ij}$ represents the performance rate of alternative $i$ ($i \epsilon \{1,2, \ldots, m\}$), with respect to criterion $j$ ($j \epsilon \{1,2, \ldots, n\}$). The normalized value of $x_{ij}$ is denoted by $\bar{x}_{ij}$.

The second step in the WASPAS technique involves determining the relative importance of the alternatives using the WSM method. This can be accomplished using Equation (22).

$$Q_i^{(1)} = \sum_{j=1}^{n} \bar{x}_{ij} w_j \qquad \forall i \qquad (22)$$

The weight of criteria $j$ ($w_j$) is determined in the previous stage using the BBWM technique. $Q_i^{(1)}$ is the total relative importance factor of alternative $i$, calculated using the WSM technique.

Similarly, in Equation (23) the total relative importance factor of the alternative $i$ is calculated using the WPM technique ($Q_i^{(2)}$).

$$Q_i^{(2)} = \prod_{j=1}^{n} (\bar{x}_{ij})^{w_j} \qquad \forall i \qquad (23)$$

Equation (24) calculates the importance factor of each alternative. Using $Q_i$, the ranking process can be done.

$$Q_i = \lambda Q_i^{(1)} + (1 - \lambda) Q_i^{(2)} \qquad \forall i \qquad (24)$$

The purpose of Equations (25-28) is to determine the optimal value of $\lambda$ by taking into account the standard deviation. It should be noted that if $\lambda = 0$, the WASPAS model reduces to the WPM model, while if $\lambda = 1$, the WSM model becomes the sole component of the WASPAS method. Equations (27-28) are used to calculate the variances.

$$\lambda = \frac{\sigma^2(Q_i^{(2)})}{\sigma^2(Q_i^{(1)}) + \sigma^2(Q_i^{(2)})} \qquad (25)$$

$$\sigma^2(Q_i^{(1)}) = \sum_{j=1}^{n} w_j^2 \sigma^2(\bar{x}_{ij}) \qquad (26)$$

$$\sigma^2(Q_i^{(2)}) = \sum_{j=1}^{n} \left( \frac{\prod_{j=1}^{n}(\bar{x}_{ij})^{w_j} w_j}{(\bar{x}_{ij})^{(w_j)}(\bar{x}_{ij})^{(1-w_j)}} \right)^2 \sigma^2(\bar{x}_{ij}) \qquad (27)$$

$$\sigma^2(\bar{x}_{ij}) = (0.05 \bar{x}_{ij})^2 \qquad (28)$$

## 2.3. Designing the Portfolio using an Optimization Model

In this step of the proposed methodology, based on the results of Sections 2.1 and 2.2, a mathematical optimization problem is developed to design the optimal portfolio for MPI's distributors.

$$Max \; Z_1 = (1 - T)R \tag{29}$$

$$Max \; Z_2 = \sum_{i=1}^{n} u_i x_i \tag{30}$$

S.t.

$$R = \sum_{i=1}^{n}(S_i - C_i)x_i \tag{31}$$

$$T = \sum_{b=1}^{m} t_b y_b \tag{32}$$

$$\sum_{b=1}^{m} y_b = 1 \tag{33}$$

$$P_{b-1} y_b - M(1 - y_b) \leq R \leq P_b y_b + M(1 - y_b) \qquad \forall b \in \{1,2,\ldots,m\}, P_0 = 0 \tag{34}$$

$$\sum_{i=1}^{n} C_i x_i \leq B \tag{35}$$

$$x_i, y_b \in \{0,1\} \qquad \forall i \in \{1,2,\ldots,n\}, b \in \{1,2,\ldots,m\} \tag{36}$$

The first objective of this model is to maximize the total profit of the MPI's distributor. This objective has been asserted mathematically in Equation (29), where $R$ shows the before-tax profit of the company, and $T$ represents the tax rate of the company.

The second objective of the proposed model, which has been illustrated in Equation (30), is to maximize the total preferability of the MPI's distributors portfolio. It should be remembered that the preferability of each project ($u_i$) has been calculated using a hybrid MADM technique discussed in Section 2.2.

The profit of the company ($R$) is calculated in Equation (31), where $S_i$ and $C_i$ show the revenue and costs of $i^{th}$ MPI's project. Also, $x_i$ is a binary variable that will be equal to 1 if the $i^{th}$ MPI's project is considered in the portfolio of the distributor, otherwise, it is equal to 0.

This model considers the progressive taxation system. In Equation (32), the tax rate of the company is calculated based on the bracket that its profit level falls in. Here, $t_b$ shows the tax rate of bracket $b$ while $y_b$ is a binary variable that will be equal to one if the profit level falls into bracket $b$ and otherwise zero. Also, Equation (35) makes sure that the revenue only falls into one of the taxation brackets.

Inequation (34) determines in which bracket the profit level of the company belongs. In this inequation, $M$ is a large number. According to this, when the profit is between $P_b$ and $P_{b-1}$, $y_b$ takes the value of one and otherwise zero.

The proposed model is constrained by inequation (35). This relation asserts that the total cost of the selected portfolio cannot be more than the distributor's available budget. It should be mentioned that $C_i$ represents the cost of project $i$.

In this mathematical model, we considered the revenue of the project in one period. If we can predict the revenue of each MPI project's revenue in each period, we should consider the inflation rate and calculate the present value of each project's revenue.

It is worth mentioning that for solving the proposed bi-objective model this research uses the Weighted Linear Combination model (Fishburn, 1967). In this technique, first, the optimal value of each objective function ($Z_i^*$) will be determined solely and without considering the rest of the objective functions. Then, instead of the existing objective functions, the combination ($Z_w$), as shown in Equation (37), will be minimized while $w \in [0,1]$.

$$Z_w = w\left(\frac{Z_1^* - Z_1}{Z_1^*}\right) + (1-w)\left(\frac{Z_2^* - Z_2}{Z_2^*}\right) \tag{37}$$

## 3. Data & Preliminary Analysis

In this section, the preliminary analysis of the gathered dataset is proposed. Firstly, Figure 4 represents the box plots of the dataset before and after classification. It clearly shows that using the classification part of the proposed methodology, the asymmetric dataset available is now three different balanced datasets.

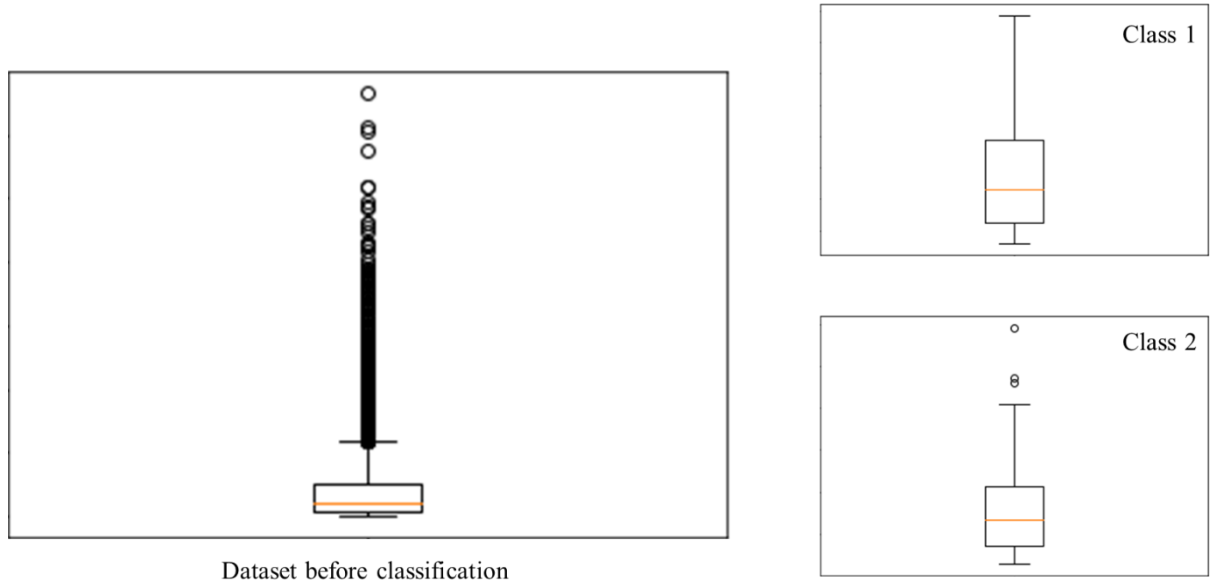

*Figure 4. Dataset's box plot before and after classification*

Furthermore, Table 5 lists the performance measures of the applied classification algorithms. Here, it is essential to remember that although some of these performance measures are designed for binary classification, the following results have been calculated using a weighted average between different classes.

*Table 5. Performance of classification algorithms.*

| Classification Algorithm | Performance Measure | | | | |
| --- | --- | --- | --- | --- | --- |
| | **Accuracy** | **Precision** | **Recall** | **F1-score** | **MCC** |
| Random Forest | 0.8677 | 0.8533 | 0.8697 | 0.8652 | 0.7332 |
| Extra Trees | 0.8490 | 0.8432 | 0.8490 | 0.8455 | 0.7340 |
| Stochastic Gradient Boosting | 0.8525 | 0.8482 | 0.8525 | 0.8501 | 0.7454 |
| XG Boost | 0.8483 | 0.8443 | 0.8483 | 0.8462 | 0.7373 |
| Light Gradient Boosting | 0.8589 | 0.8539 | 0.8589 | 0.8560 | 0.7598 |
| Cat Boost | 0.8504 | 0.8457 | 0.8504 | 0.8479 | 0.7405 |
| Decision Tree | 0.7777 | 0.7918 | 0.7777 | 0.7826 | 0.6934 |
| Support Vector Machine | 0.8497 | 0.8432 | 0.8497 | 0.8434 | 0.6899 |
| Logistic Regression | 0.8447 | 0.8382 | 0.8447 | 0.8375 | 0.7163 |
| K-Nearest Neighbors | 0.7833 | 0.7710 | 0.7833 | 0.7701 | 0.5519 |
| Voting Ensemble | 0.8699 | 0.8692 | 0.8690 | 0.8682 | 0.7502 |

Furthermore, Figure 5 displays a comparison between the accuracy performance of various algorithms during both the training and testing phases to verify that they are not over-fitted or under-fitted. The results reveal that all the classification ML algorithms utilized during training and testing demonstrate comparably high accuracy, indicating that none of them are over-fitted or under-fitted.

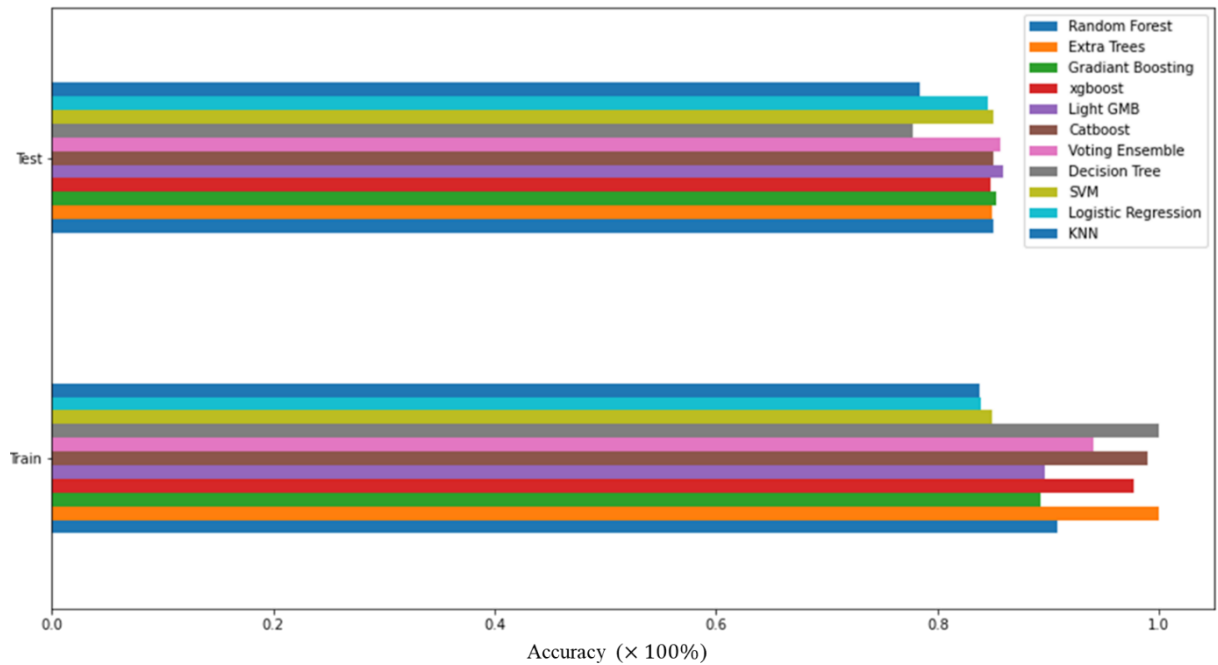

*Figure 5. Comparison of accuracy score of the applied classification algorithms in training and testing*

According to Table 5, Random Forest and Voting Ensemble algorithms have shown outstanding performance with negligible differences. However, since there is a bigger gap between the training and testing accuracy of the Voting Ensemble algorithm, the Random Forest classifier is considered the best-fitted algorithm for classifying movies.

Table 6 represents the performance of regression models in each class of movies. Based on the results, Linear Regression and Ridge Regression have the best performance for classes 1, and 2, respectively.

*Table 6. Summary of the regression models' performance in each class*

| Class | Regression Algorithm | Performance Measure | | | | |
|---|---|---|---|---|---|---|
| | | MAE | MSE | RMSE | MAPE (%) | $R^2$ |
| 1 | Linear Regression | 0.63474 | 0.61290 | 0.78288 | 1.33488 | 0.82299 |
| | Ridge Regression | 0.63484 | 0.61287 | 0.78286 | 1.33168 | 0.82312 |
| | Lasso Regression | 0.63502 | 0.61264 | 0.78271 | 1.69377 | 0.82332 |
| | Decision Tree | 0.64972 | 0.69419 | 0.83318 | 2.16483 | 0.71994 |
| | Random Forest | 0.60608 | 0.58676 | 0.76600 | 1.95999 | 0.85613 |
| | Gradient Boost | 0.60778 | 0.59655 | 0.77237 | 2.03180 | 0.84372 |
| | XG Boost | 0.59599 | 0.56837 | 0.75390 | 1.97784 | 0.87944 |
|   | Linear Regression | 0.53149 | 0.47564 | 0.68967 | 2.14764 | 0.90190 |
|   | Ridge Regression | 0.53152 | 0.47524 | 0.68938 | 2.13334 | 0.90170 |
|   | Lasso Regression | 0.53045 | 0.47436 | 0.68873 | 2.56141 | 0.90377 |
| 2 | Decision Tree | 0.56498 | 0.58547 | 0.76516 | 3.20370 | 0.74069 |
|   | Random Forest | 0.53002 | 0.47411 | 0.68856 | 2.70760 | 0.90414 |
|   | Gradient Boost | 0.54695 | 0.55769 | 0.74679 | 3.13012 | 0.78146 |
|   | XG Boost | 0.50969 | 0.47339 | 0.68803 | 2.88722 | 0.90520 |

## 4. Method Validation

To verify the proposed method in predicting the MPI projects' box offices, this section studies the gap between actual and predicted box offices for 40 movies released in 2022. This dataset, which is provided for prospective readers as Supplementary Material IV, lists the information about these movies and proposes the predicted variables (class and box office). However, to avoid any data reidentification, any non-relevant data that can be used for this reason has been eliminated.

In the classification step, 6 projects out of 40 were assigned to the wrong class (12.5% error), mostly near the borders indicated in Table 2. Also, after performing the regression step for each class the average and weighted average gap between the predicted and actual box offices were 17.07% and 15.03%.

Moreover, since employing Chat GPT as an expert is one of the main contributions of this research, it is essential to validate its compatibility with the real world. To do so, a group of 5 human experts is gathered to give their independent opinion about 100 celebrities working this year (2023). These celebrities are among different categories, including writers, actors, and directors, and also can be classified in various fame levels. They are asked to determine the fame score of each celebrity from 0 to 10. The same is asked from Chat GPT. Comparing the fame score Chat GPT provided with the average value provided by the human experts, 9.2% difference is observed which proved Chat GPT's alignment with the real-world data.

## 5. Numerical Experiments

In this section, a few test problems are investigated to check the properties of the proposed method. General Algebraic Modeling System (GAMS) software (BARON solver) has been used for solving these problems. The GAMS code can be found in Supplementary Material V.

In this research, to solve the multi-objective model proposed, the Weighted Sum Method (Athan & Papalambros, 1996) has been applied. Here it should be mentioned that for the solutions introduced for each problem, it has been assumed that the objective functions have the same weight of importance. Also, for each test problem, the Pareto front is proposed.

Throughout the following test problems, progressive taxation is considered as described in the following assumed tax brackets:

$$T = \begin{cases} 10\%, & R < 100{,}000 \\ 20\%, & 100{,}000 \leq R < 1{,}000{,}000 \\ 30\%, & 1{,}000{,}000 \leq R < 100{,}000{,}000 \\ 40\%, & R \geq 100{,}000{,}000 \end{cases} \tag{38}$$

## 5.1. Test Problem I

The first test problem under study is a small size problem consisting of only 4 projects. Table 7 lists the parameters and the results of the first test problem. The system's budget has been considered as 400,000 USD. Furthermore, Figure 6 represents the Pareto front for this problem. After solving the problem, the values of the first and second objective functions have been calculated as 7,611,101.4 and 0.418, respectively (considering $w = 0.5$ in Equation (37)). Also, the total profit of the company before taxation was calculated as 10,873,000, and the total tax paid was calculated to be 3,261,900.6.

*Table 7. Parameters and the final solution of Test Problem I.*

| $i$ | $S_i$ | $U_i$ | $C_i$ | $x_i$ |
|---|---|---|---|---|
| 1 | 567,328 | 0.193 | 123,562 | 0 |
| 2 | 2,574,291 | 0.307 | 158,976 | 1 |
| 3 | 8,664,518 | 0.111 | 206,831 | 1 |
| 4 | 2,345,183 | 0.389 | 215,123 | 0 |

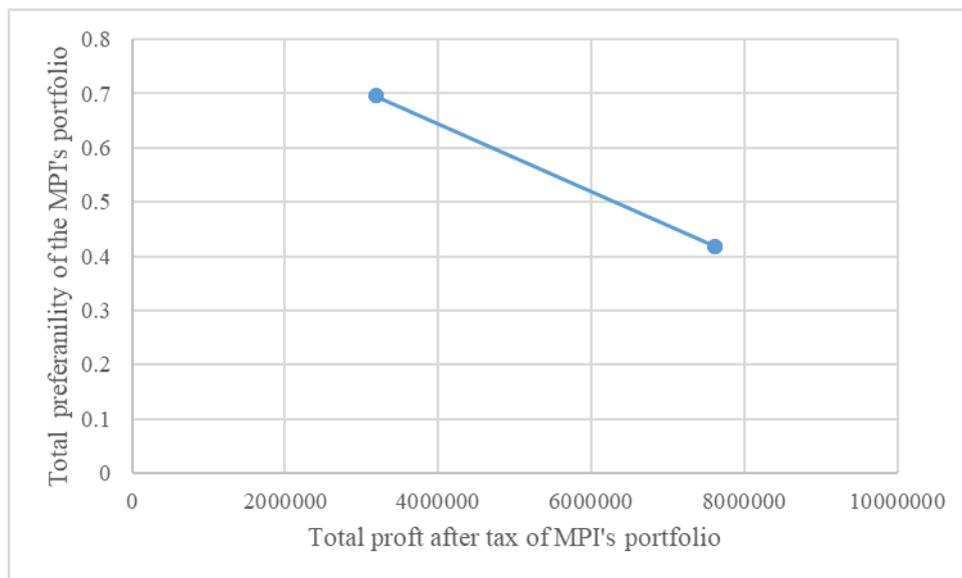

*Figure 6. Pareto front of test problem I.*

## 5.2. Test Problem II

The second examination scenario under investigation pertains to an intermediate-scale problem comprising ten distinct projects. The parameters and outcomes for the initial test problem are documented in Table 8 for reference. The financial allocation for the system is firmly established at 800,000 USD. Additionally, Figure 7 graphically illustrates the Pareto front applicable to this particular problem. Following the problem-solving process, the computed values for the primary and secondary objective functions are determined to be 51,464,520 and 0.342, respectively. (considering $w = 0.5$ in Equation (37)). Also, the total profit of the company before taxation was calculated as 73,520,740, and the total tax paid was calculated to be 22,056,220.

*Table 8. Parameters and the final solution of Test Problem II.*

| $i$ | $S_i$ | $U_i$ | $C_i$ | $x_i$ |
|---|---|---|---|---|
| 1 | 3,526,133 | 0.079962875 | 396,184 | 0 |
| 2 | 16,121,814 | 0.077660488 | 201,057 | 0 |
| 3 | 4,560,689 | 0.117065546 | 384,420 | 0 |
| 4 | 22,810,973 | 0.151242461 | 386,741 | 0 |
| 5 | 22,333,171 | 0.107820196 | 230,674 | 1 |
| 6 | 25,988,901 | 0.131829017 | 218,304 | 1 |
| 7 | 7,056,443 | 0.068473314 | 288,491 | 0 |
| 8 | 26,375,329 | 0.073415951 | 233,148 | 0 |
| 9 | 25,979,138 | 0.102333847 | 331,490 | 1 |
| 10 | 10,782,147 | 0.090196305 | 264,913 | 0 |

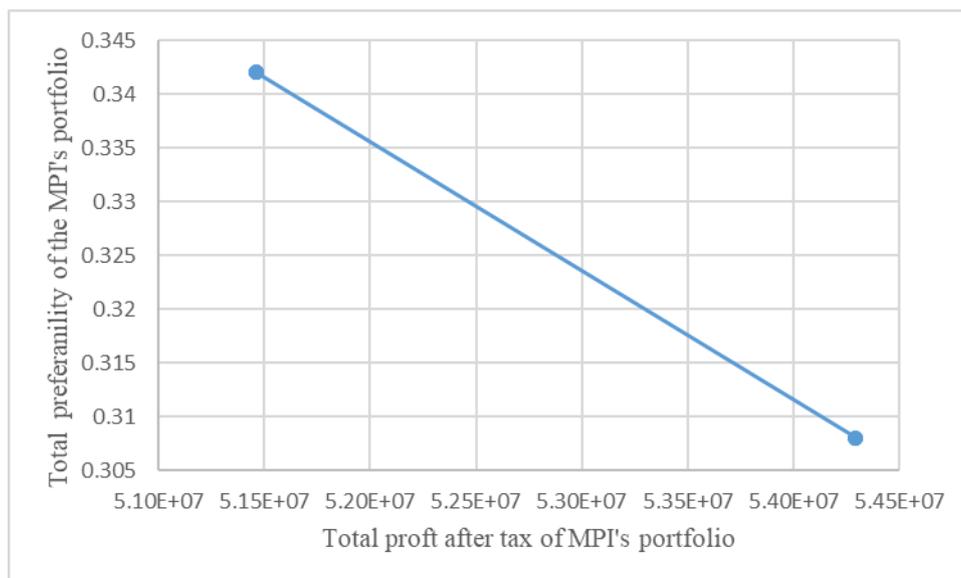

*Figure 7. Pareto front of test problem II.*

*5.3. Test Problem III*

The third test case under investigation pertains to a problem of considerable magnitude, encompassing a portfolio of 20 distinct projects. Comprehensive details regarding the parameters and outcomes of this test problem are systematically cataloged in Table 9. The financial allocation for the system is rigorously established at 1,500,000 USD. Moreover, Figure 8 serves as an illustrative representation of the Pareto front that characterizes the specific attributes of this problem. Upon the successful resolution of the problem, the computed values for the primary and secondary objective functions stand at 89,485,520 and 0.249, respectively. The total profit before tax and the total tax have been calculated as 149,142,500 and 59,657,020.

*Table 9. Parameters and the final solution of Test Problem III.*

| $i$ | $S_i$ | $U_i$ | $C_i$ | $x_i$ |
|---|---|---|---|---|
| 1 | 33,945,065 | 0.043055 | 524,557 | 0 |
| 2 | 21,855,988 | 0.035665 | 290,981 | 1 |
| 3 | 46,131,546 | 0.066716 | 525,804 | 0 |
| 4 | 21,152,250 | 0.034135 | 372,937 | 0 |
| 5 | 23,352,095 | 0.064202 | 461,080 | 0 |
| 6 | 27,023,351 | 0.060216 | 297,466 | 1 |
| 7 | 42,635,255 | 0.035325 | 357,955 | 1 |
| 8 | 11,855,641 | 0.045057 | 509,830 | 0 |
| 9 | 35,429,283 | 0.044676 | 577,567 | 0 |
| 10 | 22,782,789 | 0.043415 | 426,619 | 0 |
| 11 | 21,681,939 | 0.052839 | 585,788 | 0 |
| 12 | 25,056,002 | 0.054133 | 367,881 | 0 |
| 13 | 33,636,719 | 0.052944 | 475,728 | 0 |
| 14 | 26,288,485 | 0.047984 | 390,706 | 0 |
| 15 | 9,368,224 | 0.037523 | 289,676 | 0 |
| 16 | 12,312,749 | 0.049124 | 331,457 | 0 |
| 17 | 13,604,840 | 0.057295 | 516,763 | 0 |
| 18 | 42,091,295 | 0.062121 | 321,139 | 1 |
| 19 | 17,027,318 | 0.05562 | 223,126 | 1 |
| 20 | 33,278,343 | 0.057954 | 385,842 | 0 |

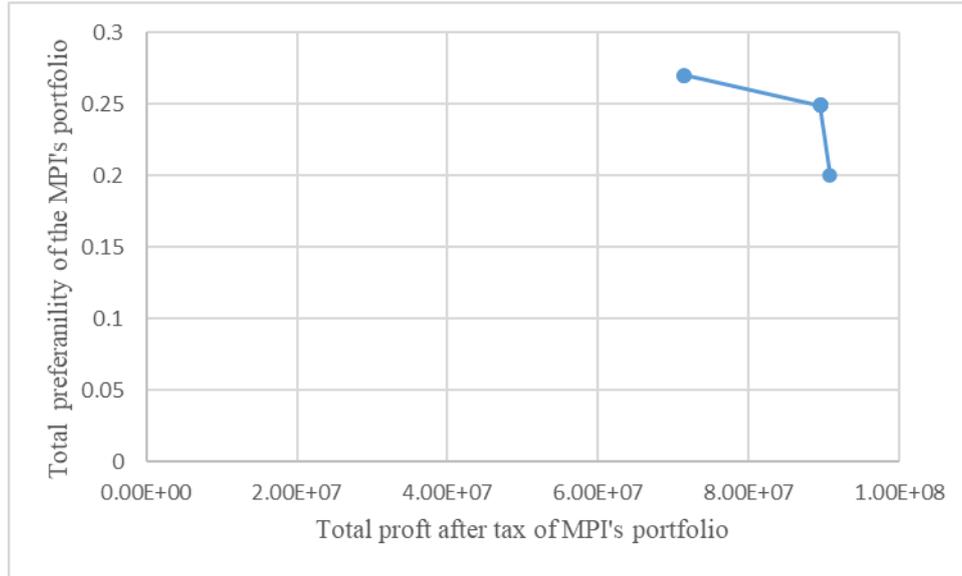

*Figure 8. Pareto front of test problem III.*

## 6. Sensitivity Analysis

### 6.1. Fame Score Features

To check the influence of the three fame score features (Director's Fame Score, Writer's Fame Score, Lead Actor/Actress Fame Score) determined utilizing Chat GPT, the ML models are once applied while these three features have been dropped. Table 10 shows the results of regression after this change.

*Table 10. Results of the regression algorithms removing fame scores*

| Class | Regression Algorithm | Performance Measure | | | | |
|---|---|---|---|---|---|---|
| | | MAE | MSE | RMSE | MAPE (%) | $R^2$ |
| 1 | Linear Regression | 7.17257 | 5.57739 | 2.36165 | 10.81253 | 0.20575 |
| | Ridge Regression | 7.11021 | 5.63841 | 2.37454 | 10.91978 | 0.20578 |
| | Lasso Regression | 7.23923 | 5.69756 | 2.38696 | 14.05830 | 0.20583 |
| | Decision Tree | 6.62715 | 5.55352 | 2.3566 | 15.15381 | 0.21598 |
| | Random Forest | 5.57594 | 4.10732 | 2.02666 | 9.79995 | 0.25684 |
| | Gradient Boost | 5.83469 | 5.36895 | 2.3171 | 10.15900 | 0.25312 |
| | XG Boost | 5.90031 | 5.11533 | 2.26171 | 11.86704 | 0.26383 |
| 2 | Linear Regression | 6.00584 | 4.32833 | 2.08047 | 8.80533 | 0.22548 |
| | Ridge Regression | 5.95303 | 4.37221 | 2.09099 | 8.96003 | 0.22543 |
| | Lasso Regression | 6.04713 | 4.41155 | 2.10037 | 11.01407 | 0.22594 |
| | Decision Tree | 5.76280 | 4.68376 | 2.1642 | 9.61110 | 0.22221 |
| | Random Forest | 4.87619 | 3.31877 | 1.82175 | 8.12280 | 0.27124 |
| | Gradient Boost | 5.41481 | 5.01921 | 2.24036 | 9.39036 | 0.23444 |

| | XG Boost | 4.89303 | 4.26051 | 2.06411 | 8.66166 | 0.27156 |

Comparing Table 6 and Table 10, it is clear that the performance of the regression algorithms decreases drastically. However, tree-based algorithms, especially random forest algorithm, seem to perform more robustly to feature removal.

*6.2. Classification Step*

This section is dedicated to studying the effect of benefitting from the classification step in the methodology. Table 11 lists the regression algorithms' results without classifying the dataset first. That is to say, after preprocessing and eliminating the outlier data (previously introduced as class 3), we applied the regression algorithms to measure their performance.

*Table 11. Performance of regression algorithms without classification of the dataset*

| Regression Algorithm | Performance Measure | | | | |
|---|---|---|---|---|---|
| | MAE | MSE | RMSE | MAPE (%) | $R^2$ |
| **Linear Regression** | 30.1788 | 20.4645 | 4.52377 | 128.48246 | 0.037723 |
| **Ridge Regression** | 30.1934 | 20.4576 | 4.52301 | 127.37860 | 0.038911 |
| **Lasso Regression** | 30.1689 | 20.4616 | 4.52344 | 167.03873 | 0.037732 |
| **Decision Tree** | 28.8483 | 20.9794 | 4.58032 | 193.05716 | 0.036156 |
| **Random Forest** | 26.7880 | 17.1739 | 4.14413 | 183.84543 | 0.047737 |
| **Gradient Boost** | 26.7783 | 17.6919 | 4.20618 | 192.36251 | 0.046160 |
| **XG Boost** | 26.8154 | 17.9554 | 4.23739 | 189.64094 | 0.045358 |

Just by eyeballing the difference between Table 6 and Table 11, the necessity of applying classification in the dataset is clear.

*6.3. Optimization Model*

To understand the behavior of the proposed optimization model, the effect of changing in budget on the profit of the company is investigated here. For this, the budget of the third test problem ($w$ is set on 0.5) is changed from 0 to 3,750,000 with a step length of 250,000. Figure 9 shows the behavior of $Z_1$, $R$, and total tax in each budget.

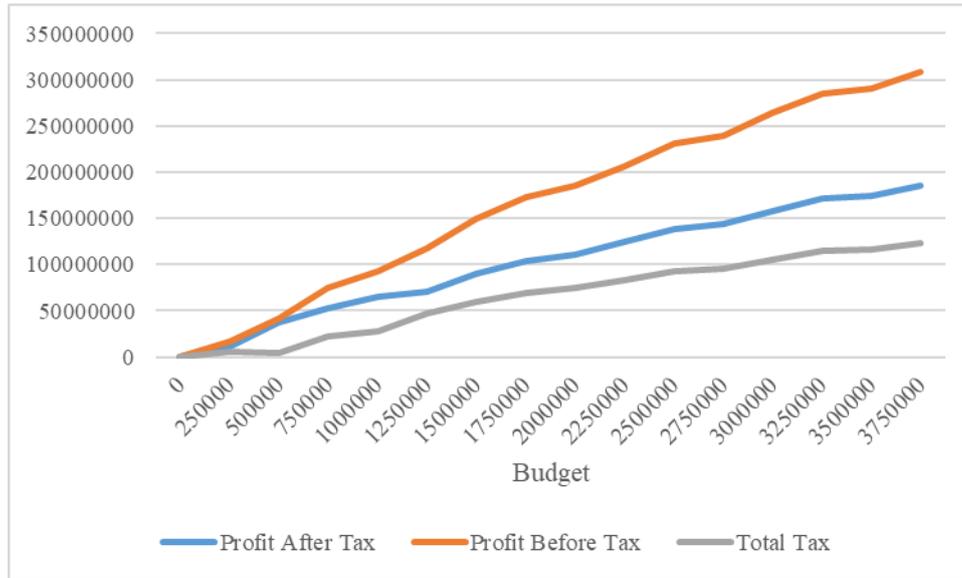

*Figure 9. Effect of budget on $Z_1$, R, and total tax*

According to Figure 9, as the budget increases, both profits before and after tax also increase. This suggests that the distribution company is likely to see a return on its investments, as increased spending correlates with higher profits.

Also, Table 12 lists the projects selected as a part of the distributor's portfolio at each level of budget.

*Table 12. Effect of budget on portfolio's project selection*

| Budget | \multicolumn{20}{c}{Projects} |
|---|---|---|---|---|---|---|---|---|---|---|---|---|---|---|---|---|---|---|---|---|
|  | 1 | 2 | 3 | 4 | 5 | 6 | 7 | 8 | 9 | 10 | 11 | 12 | 13 | 14 | 15 | 16 | 17 | 18 | 19 | 20 |
| 0 |  |  |  |  |  |  |  |  |  |  |  |  |  |  |  |  |  |  |  |  |
| 250000 |  |  |  |  |  |  |  |  |  |  |  |  |  |  |  |  |  |  | ● |  |
| 500000 |  |  |  |  |  |  |  |  |  |  |  |  |  |  |  |  |  | ● |  |  |
| 750000 |  |  |  |  |  |  |  |  |  |  |  |  |  |  |  |  |  | ● |  | ● |
| 1000000 |  |  |  |  |  | ● |  |  |  |  |  | ● |  |  |  |  |  | ● |  |  |
| 1250000 |  |  |  |  |  | ● |  |  |  |  |  |  |  |  |  |  |  | ● | ● | ● |
| 1500000 |  | ● |  |  |  | ● | ● |  |  |  |  |  |  |  |  |  |  | ● | ● |  |
| 1750000 |  |  | ● |  |  | ● | ● |  |  |  |  |  |  |  |  |  |  | ● | ● |  |
| 2000000 |  |  |  |  |  | ● | ● |  |  |  |  | ● |  |  |  |  |  | ● | ● | ● |
| 2250000 |  | ● |  |  |  | ● | ● |  |  |  |  | ● |  |  |  |  |  | ● | ● | ● |
| 2500000 |  |  | ● |  |  | ● | ● |  |  |  |  | ● |  |  |  |  |  | ● | ● | ● |
| 2750000 |  | ● | ● |  |  | ● | ● |  |  |  |  | ● |  |  | ● |  |  | ● | ● | ● |
| 3000000 |  |  | ● |  |  | ● | ● |  |  |  |  | ● | ● |  |  |  |  | ● | ● | ● |
| 3250000 |  | ● | ● |  |  | ● | ● |  |  |  |  | ● | ● |  |  |  |  | ● | ● | ● |
| 3500000 |  | ● | ● |  |  | ● | ● |  |  |  |  | ● |  | ● |  | ● |  | ● | ● | ● |

| 3750000 | • | • | • | • | | • | • | | • | • | • |

## 7. Conclusion

In conclusion, this study has addressed a persistent challenge within the MPI concerning portfolio management, which has hitherto remained largely unaddressed. To formulate an optimal portfolio strategy for MPI distributors, it is imperative to make accurate predictions regarding the box office performance of individual projects. Significantly, this research has introduced an innovative approach by incorporating the influence of celebrities associated with each MPI project, a dimension largely overlooked in prior expert-based methodologies.

In this paper, For the first time in decision science history, an LLM is benefitted from as an expert. The verification conducted suggests the reliability of the proposed method for predicting the box office.

The research methodology can be summarized as follows: the fame scores of celebrities are quantified utilizing Chat GPT, subsequent classification of projects is performed to mitigate data imbalances inherent to MPI datasets, and precise box office predictions are generated for projects within each classification. Furthermore, through the application of a hybrid MADM technique, the desirability of each project from the distributor's perspective is quantified. Ultimately, by leveraging a bi-objective optimization model, this study culminates in the formulation of an optimal portfolio tailored to meet the unique needs and preferences of MPI distributors. This innovative approach contributes to the advancement of portfolio management practices within the MPI, offering a valuable framework for strategic decision-making in this dynamic sector.

Although the authors did their best to make this methodology as comprehensive as possible, there are a few limitations:

- To understand the application of the proposed study, it is essential to employ it in the real world for at least one MPI distributor. However, it was not possible in the timeline of this research.
- The dataset used for training and testing (randomly selected) contains a part of the movies released in the period under study. Using a more comprehensive data set, the results would probably be more accurate.

Finally, here the authors would like to propose a few suggestions for future research:

- Investigate the role of other factors that may have an impact on a movie's box office, including inflation rate, unemployment rate, and Gross Domestic Product per capita in that year.

- In this research, it has been assumed that all parameters and variables are crisp and deterministic. We suggest in future research uncertainty (fuzzy and/or stochastic) be considered.
- This research benefited from one of the LLMs (Chat GPT). For future research, it is rational to compare the responses of different models.
- The proposed technique can be used for portfolio optimization in various industries. We recommend future research to apply this methodology and check its properties in different industries.